\documentclass[a4paper, 10pt, conference]{ieeeconf}      % Use this line for a4
\usepackage{FG2021}

\usepackage{graphicx}
\usepackage{subfig}
\usepackage{multirow}
\usepackage{makecell}
\usepackage{url} % Add this line

\graphicspath{ {./rai_aria_images/} }

\FGfinalcopy % *** Uncomment this line for the final submission

\IEEEoverridecommandlockouts                              % This command is only
\title{\LARGE \bf
EgoBlur: Responsible Innovation in Aria
}

\author{\parbox{16cm}{\centering
    {\large Nikhil Raina, Guruprasad Somasundaram, Kang Zheng, Sagar Miglani, Steve Saarinen, Jeff Meissner, Mark Schwesinger, Luis Pesqueira, Ishita Prasad, Edward Miller, Prince Gupta, Mingfei Yan, Richard Newcombe, Carl Ren, Omkar M Parkhi}\\
    {\normalsize
     Meta Reality Labs
    }\\
    {\small
     \{nrraina, guruprasad, zhengkang, sagarmiglani, saarinen, jeffme, schwes, lpesqueira, isprasad, edwardmiller, guptaprince, mingfeiy, newcombe, carlren, omkar\}@meta.com
    }
}
}

\begin{document}

\ifFGfinal
\thispagestyle{empty}
\pagestyle{empty}
\else
\author{Anonymous FG2021 submission\\ Paper ID \FGPaperID \\}
\pagestyle{plain}
\fi
\maketitle

%%%%%%%%%%%%%%%%%%%%%%%%%%%%%%%%%%%%%%%%%%%%%%%%%%%%%%%%%%%%%%%%%%%%%%%%%%%%%%%%
\begin{abstract}

Project Aria pushes the frontiers of Egocentric AI with large-scale real-world data collection using purposely designed glasses with privacy first approach. To protect the privacy of bystanders being recorded by the glasses, our research protocols are designed to ensure recorded video is processed by an AI anonymization model that removes bystander faces and vehicle license plates.  Detected face and license plate regions are processed with a Gaussian blur such that these personal identification information (PII) regions are obscured. This process helps to ensure that anonymized versions of the video is retained for research purposes. In Project Aria, we have developed a state-of-the-art anonymization system~\lq EgoBlur\rq. In this paper, we present extensive analysis of EgoBlur %our face and license plate anonymization system % 
on challenging datasets comparing its performance with other state-of-the-art systems from industry and academia including extensive Responsible AI analysis on recently released Casual Conversations V2~\cite{porgali2023casual} dataset.

\end{abstract}

%%%%%%%%%%%%%%%%%%%%%%%%%%%%%%%%%%%%%%%%%%%%%%%%%%%%%%%%%%%%%%%%%%%%%%%%%%%%%%%%
\section{INTRODUCTION}

As part of our commitment to a privacy-first approach in Project Aria, we are committed to anonymizing people's faces and vehicle license plates in our recordings. Our anonymization system operates within our data ingestion platform, ensuring that human faces and license plates are obfuscated before a recording is made available for research purposes. We conducted extensive evaluations of our system on various challenging datasets across different axes of evaluation. Additionally, we performed a detailed Responsible AI analysis of our face detection model. The goal of this paper is to present the results of that analysis and compare it to a few other state-of-the-art systems.

The objective of EgoBlur is to obscure human faces and vehicle license plates captured by Aria glasses. While there have been previous works on in-place face editing and replacement, which could serve as obfuscation strategies~\cite{klemp2023ldfa, balaji2021temporally,ma2021cfa}, these methods have not been extensively tested on real-world videos from a user's egocentric perspective. We opt for a simpler yet effective approach of detecting these objects (faces and license plates) using traditional object detectors and obfuscating the underlying pixels with a Gaussian blur function. This selection subsequently opens up choices in the object detection world with several research works showcasing state-of-the-art performance on challenging datasets for face detection as well as for generic object detection.

We select FasterRCNN~\cite{ren2015faster} as our choice of object detector system. It offers several advantages. FasterRCNN and its subsequent variants such as MaskRCNN~\cite{li2022exploring} are one of the top performing methods on benchmark datasets such as MS-COCO~\cite{lin2014microsoft}. They have been widely studied, cited, and put into production systems. They are applicable to a wide variety of objects and do not need task-specific treatment for specific classes such as facial keypoints annotations for better face detection performance.
To demonstrate the effectiveness of our choice of FasterRCNN-based generic object detector, we compare its performance for the problem of face detection/anonymization with state-of-the-art RetinaFace\cite{deng2020retinaface} and MediaPipe\cite{bazarevsky2019blazeface} face detectors. We demonstrate that our choice of using a task-agnostic detector for both face and license plate detection outperforms or matches the performance of leading techniques, achieving over 90\% recall on challenging benchmark datasets. Our anonymization pipeline is designed to be flexible, allowing for easy replacement of the underlying detector with improved versions of our detectors or any new detection model. 
In this paper, we provide details of two subsystems of EgoBlur: first, we discuss our face anonymization method, providing details on detector training and a detailed analysis of its performance compared to other state-of-the-art methods on challenging datasets. Then, we provide similar insights into the performance of our license plate anonymization method, discussing its training and analysis on our benchmarking dataset.

\section{Anonymization Benchmarking}

In this section, we provide a comprehensive overview of face and license plate anonymization subsystems. We begin by describing our training methodology briefly. We then follow it up with detailed performance analysis of the underlying detectors. 

\subsection{Faces}

% For training the FasterRCNN based face detector, we follow a weakly supervised approach~\cite{beyer2022knowledge}. More specifically, we selected a large corpus of images and used the publicly available RetinaFace model as a strong teacher to provide the pseudo ground truth. We then piped this data through the standard ResNext-101-32x8 FPN based FasterRCNN model using Detectron2~\cite{wu2019detectron2}. We followed learning rate schedule as per the long term training schedule experiments and added training data augmentation increase the share of grayscale images during training of the detector. In the following sections, we describe the datasets used for the evaluations and give detailed analysis of our evaluation.

For training the FasterRCNN-based face detector, we adopt a weakly supervised approach~\cite{beyer2022knowledge}. We selected a large corpus of images and use the publicly available RetinaFace model as a strong teacher to provide pseudo ground truth. We then feed this data through the standard ResNext-101-32x8 FPN-based FasterRCNN model using Detectron2~\cite{wu2019detectron2}. To improve performance, we follow a learning rate schedule based on long-term training experiments and increase the share of grayscale images during training. In the following sections, we describe the datasets used for evaluation and present a detailed analysis of our results.

\subsubsection{Benchmarking Datasets}

\paragraph{CCV2 Dataset}: 
The recently released Casual Conversations V2 dataset~\cite{porgali2023casual} provides valuable annotations for evaluating the performance of a model on various Responsible AI attributes, such as age, skin tone, gender, and country of origin. To leverage this dataset for our face detection benchmarking, we augmented it with manually annotated face bounding boxes. This allowed us to carefully evaluate the performance of our face detector on these important responsible AI attributes provided by the dataset. Specifically, we uniformly sampled frames from the videos of CCV2 and manually annotated them with face bounding boxes, resulting in a dataset of 259,656 bounding boxes.

%Recently released Casual Conversations V2 dataset~\cite{porgali2023casual}  provides rich annotations to evaluate performance of a model on several Responsible AI attributes such as age, skin-tone, gender, country of the actors present in the videos. To make use of this dataset for our face detection benchmarking, we augmented the dataset with manually annotated face bounding boxes.  This enabled us to carefully evaluate the performance of our face detector on these important responsible AI attributes provided by this dataset. Specifically, we uniformly sampled frames from the videos of CCV2 dataset and manually annotated frames with the face bounding boxes. We annotated 259656 frames from the CCV2 dataset creating a dataset of 259,656 bounding boxes. 

\paragraph{Aria Pilot Dataset}
The Aria Pilot Dataset~\cite{AriaPilot} is an open-source egocentric dataset collected using the Aria glasses. To use it for face detector benchmarking, we comprehensively annotated this dataset with manual face bounding box annotations. We created a dataset of 18,508 annotated frames with 23,242 bounding boxes. This complementary dataset to CCV2 provides essential in-domain data specific to the use-cases typically observed in our recordings. In addition to the bounding boxes, we augmented this dataset with various attribute labels such as wearing glasses, truncated and occluded faces, dark lighting scenarios, etc., to understand the fine-grained performance of our system. To avoid annotator bias, these annotations were carried out in a multi-review process (3 annotators labeling attributes for the same bounding box), and the attribute labels were selected using majority voting. The resultant dataset provides a strong benchmark to evaluate detection performance in common scenarios observed in our recordings and provides insights for areas of further improvement.

\subsubsection{Evaluation}

For evaluating our detectors, we use standard object detection evaluation metrics. We compute the intersection over union (IoU) with an overlap threshold of 0.5 and calculate the average precision (AP) and average recall (AR) using the MS-COCO API\cite{lin2014microsoft}. To provide a comprehensive comparison, we benchmark our detector against two publicly available face detectors: RetinaFace\cite{deng2020retinaface} and MediaPipe~\cite{bazarevsky2019blazeface}. While MediaPipe is designed for low latency applications, RetinaFace is a strong academic baseline that has demonstrated state-of-the-art results on various face detection tasks. By comparing our performance to these leading methods on the carefully annotated datasets described above, we can contextualize our results and provide a more meaningful assessment of our detector's capabilities.

%For the evaluation of our detectors, we use standard object detection evaluation metrics. We use intersection over union with an overlap threshold of 0.5 and compute the average precision (AP) and average recall (AR) using the MS-COCO api~\cite{lin2014microsoft}. As stated previously, we compare performance of our detector to  two publicly available face detectors, RetinaFace~\cite{deng2020retinaface} and MediaPipe~\cite{bazarevsky2019blazeface}. MediaPipe detector is mainly designed for low latency applications while the RetinaFace detector is a strong academic baseline showing state of the art results on various face detection tasks. Comparing with these two systems enables us to ground our results with performance of leading methods on carefully annotated datasets described above.

\begin{figure*}
\begin{tabular}{cccc}
\subfloat[]{\includegraphics[height= 2.5in, width = 1.5in]{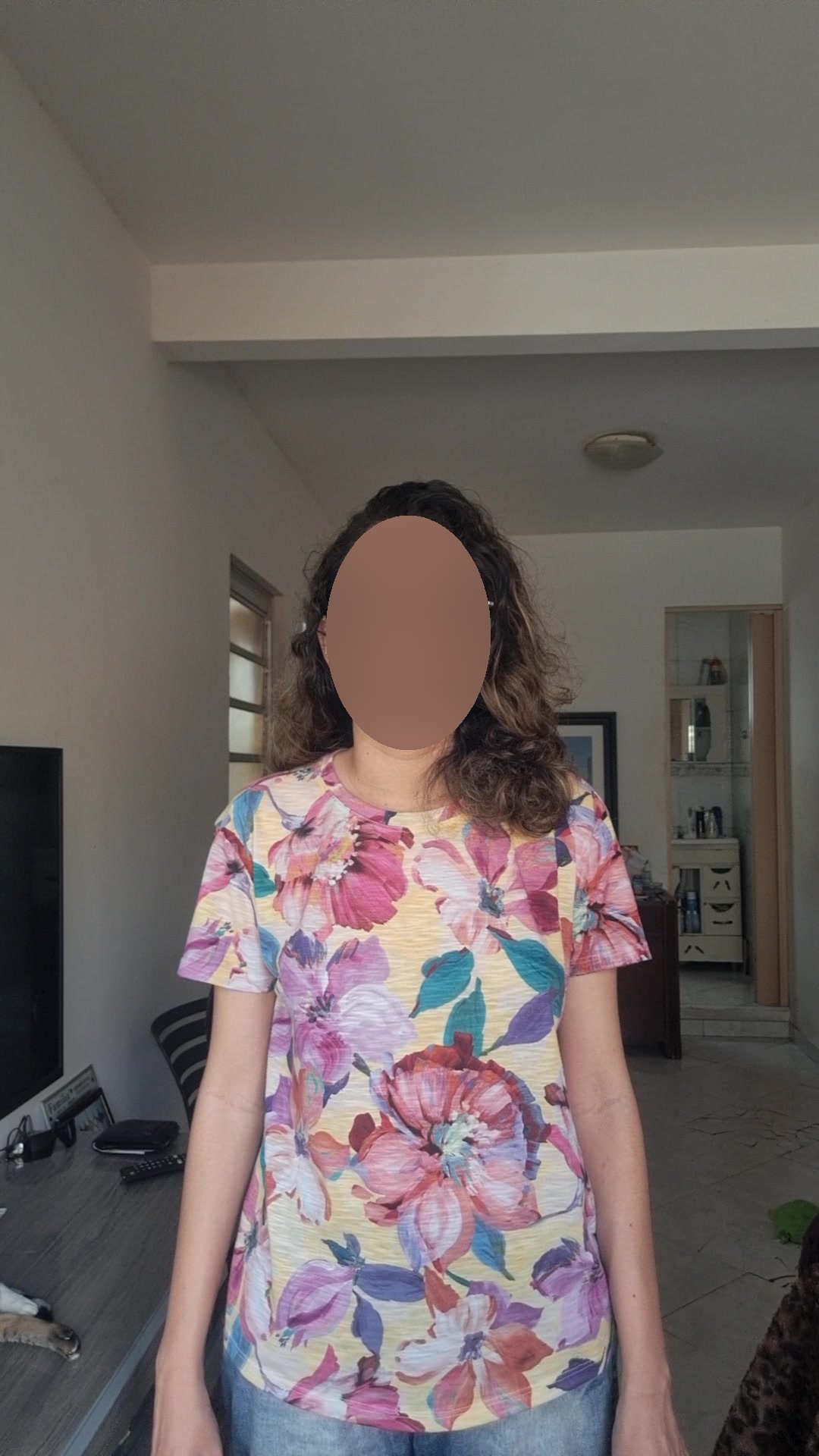}} &
\subfloat[]{\includegraphics[height= 2.5in, width = 1.5in]{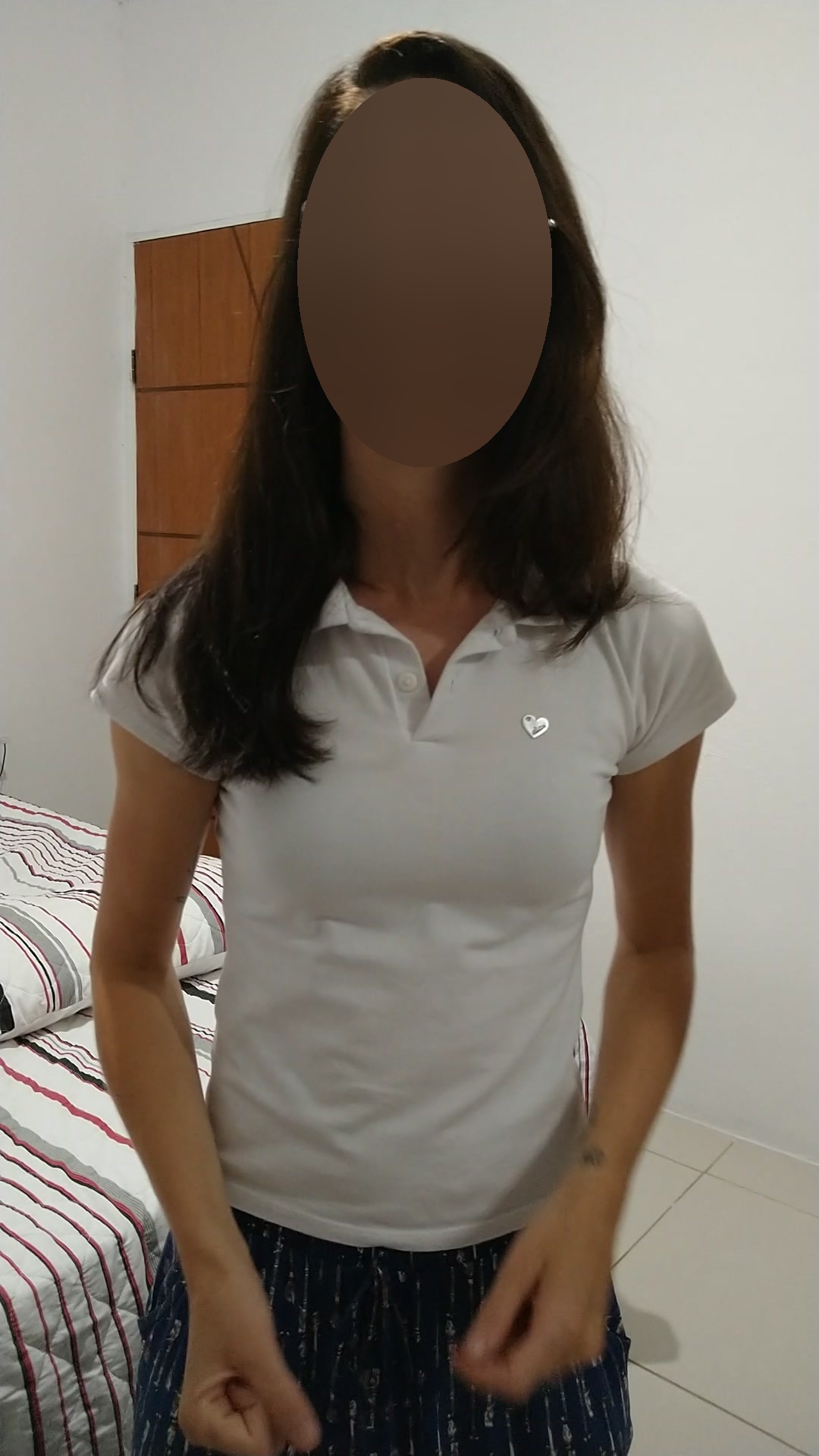}} &
\subfloat[]{\includegraphics[height= 2.5in, width = 1.5in]{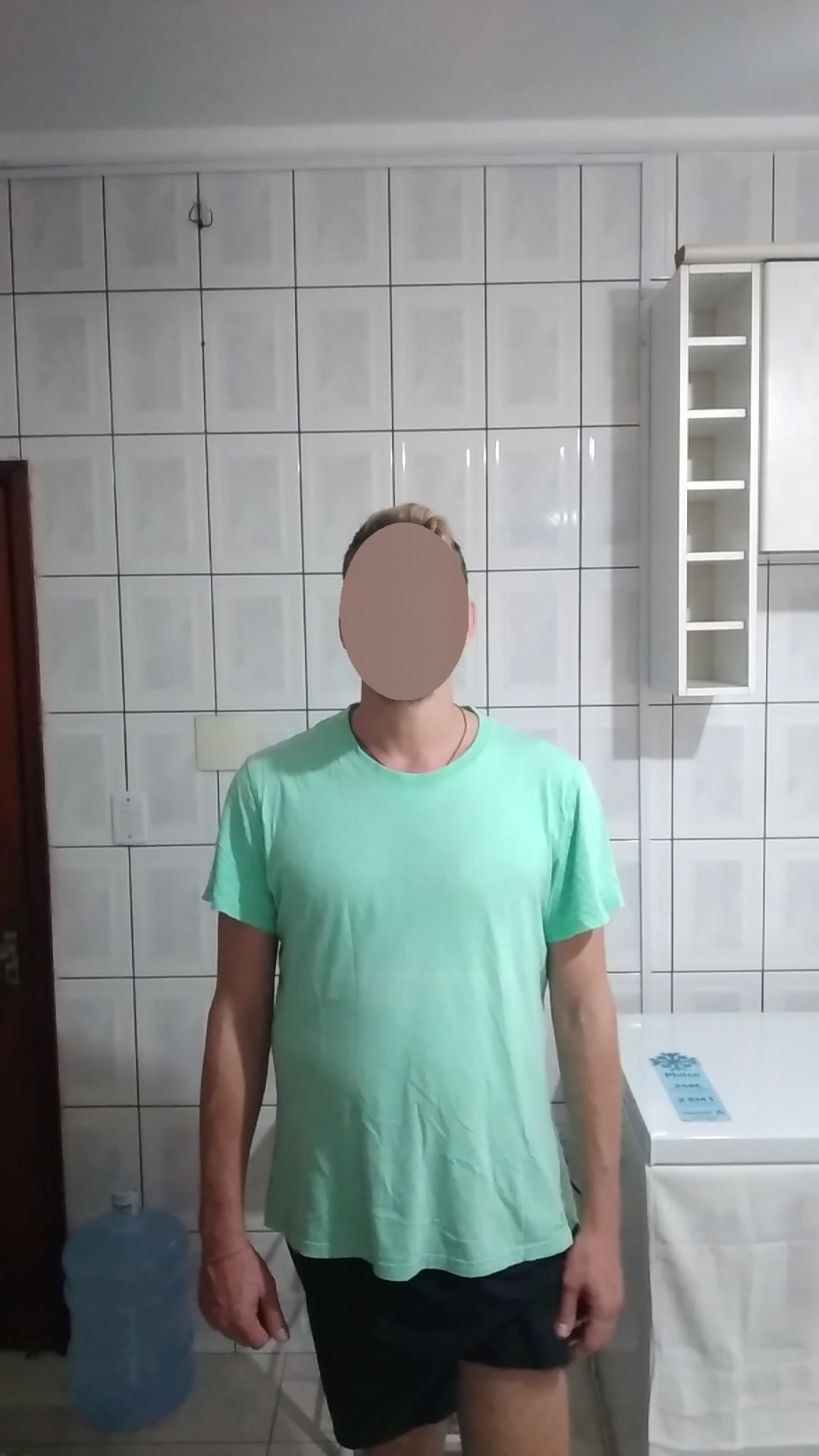}} &
\subfloat[]{\includegraphics[height= 2.5in, width = 1.5in]{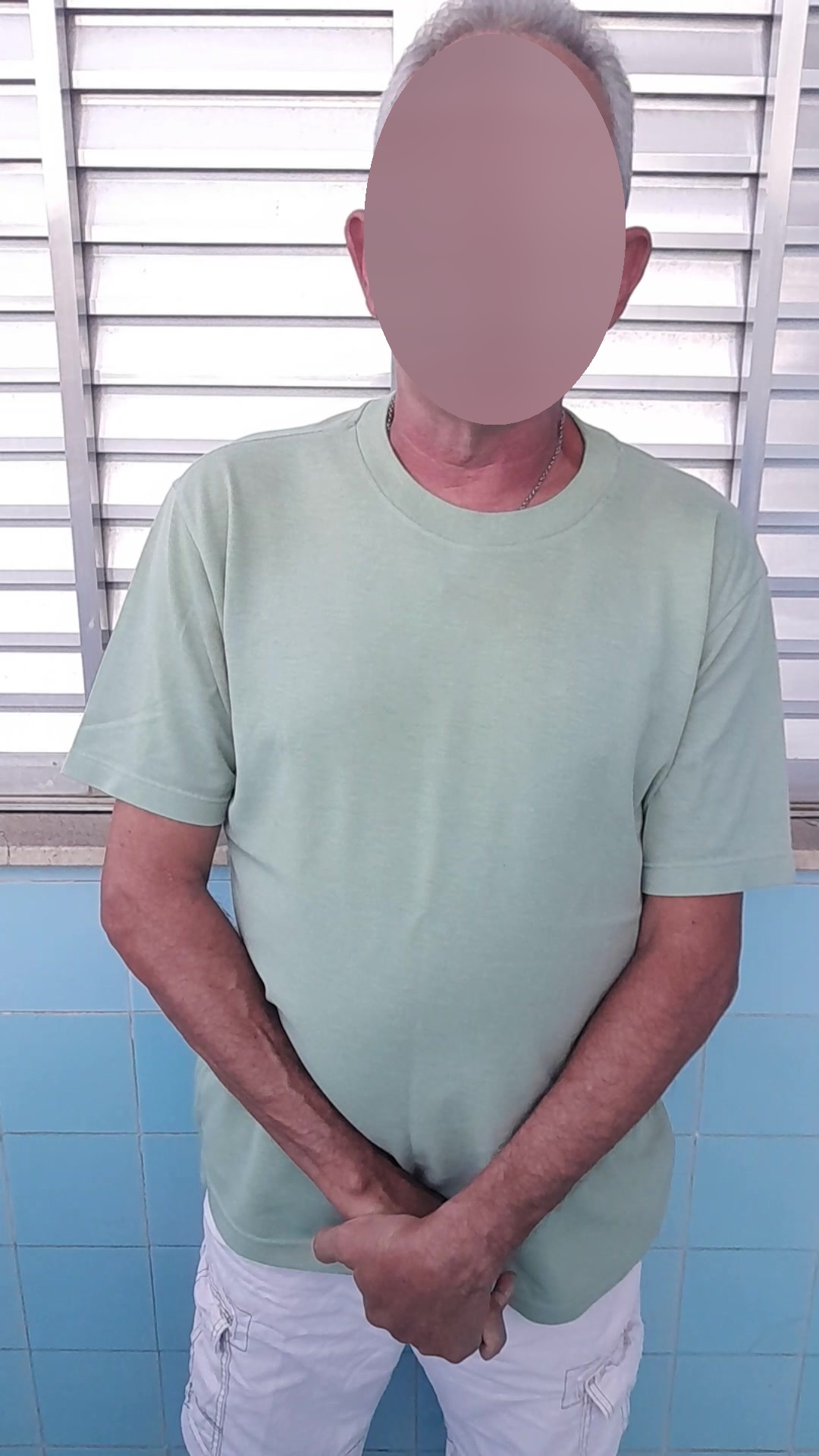}}

\end{tabular}
\caption{Qualitative results on CCV2 dataset. CCV2 dataset has actors from various countries, age group, gender and skin tone buckets. Our method provides consistent results across all buckets.}
\label{fig:ccv2_anon}

\end{figure*}

Tables~\ref{tab:ccv2_overall_eval}-\ref{tab:ccv2_country_eval} present the performance of our method on the challenging CCV2 dataset. Table~\ref{tab:ccv2_overall_eval} displays the aggregated results for the overall dataset, where our method outperforms MediaPipe and matches the performance of RetinaFace. Tables~\ref{tab:ccv2_age_eval}-\ref{tab:ccv2_country_eval} provide a detailed analysis of the performance across various attributes. Two key observations can be made from these results. Firstly, our method consistently performs better than or equal to the state-of-the-art methods. Secondly, the performance of our method is consistent across all Responsible AI buckets. Figure~\ref{fig:ccv2_anon} showcases some qualitative results of our method on the CCV2 dataset.

%Tables~\ref{tab:ccv2_overall_eval}-\ref{tab:ccv2_country_eval} show performance of our method on the challenging CCV2 dataset. Table~\ref{tab:ccv2_overall_eval} shows results aggregated on the overall dataset. Our method performs better than the MediaPipe and matches the performance of RetinaFace. Tables~\ref{tab:ccv2_age_eval}-~\ref{tab:ccv2_country_eval} show the detailed performance analysis on various attributes. There are two critical observations to be made here. First, our method consistently performs better than or equal to the performance shown by the state of the art methods. Second, the performance of our method is consistent across all the Responsible AI buckets. Figure~\ref{fig:ccv2_anon} shows few qualitative results of our method on the CCV2 dataset.

Tables~\ref{tab:aria_face_overall_eval}-\ref{tab:aria_face_grayscale_eval} present the performance of our system on the Aria Pilot Dataset. Table~\ref{tab:aria_face_overall_eval} displays the overall performance of our method compared to two baselines, where our method outperforms both and achieves a recall of over 90\%. The Aria Pilot dataset contains recordings from three different camera streams, one colored (RGB) and two gray-scale. It is noteworthy that our method's performance on both RGB and gray-scale streams is comparable, while the baseline systems exhibit a significant difference in performance across the streams.

%Tables~\ref{tab:aria_face_overall_eval}-\ref{tab:aria_face_grayscale_eval} show the performance of our system on Aria Pilot Dataset. Table~\ref{tab:aria_face_overall_eval} shows the overall performance of our method compared with other two baselines. Our method performs better than the compared baselines and manages to achieve Recall of over 90\%. The Aria Pilot dataset has recordings coming from three different camera streams, one colored and two in gray-scale. It can be seen that our method's performance on both colored (RGB) and gray-scale streams is comparable while the baseline systems show a significant difference in performance across the streams.  

\begin{figure*}
\begin{tabular}{cccc}
\subfloat[]{\includegraphics[width = 1.5in]{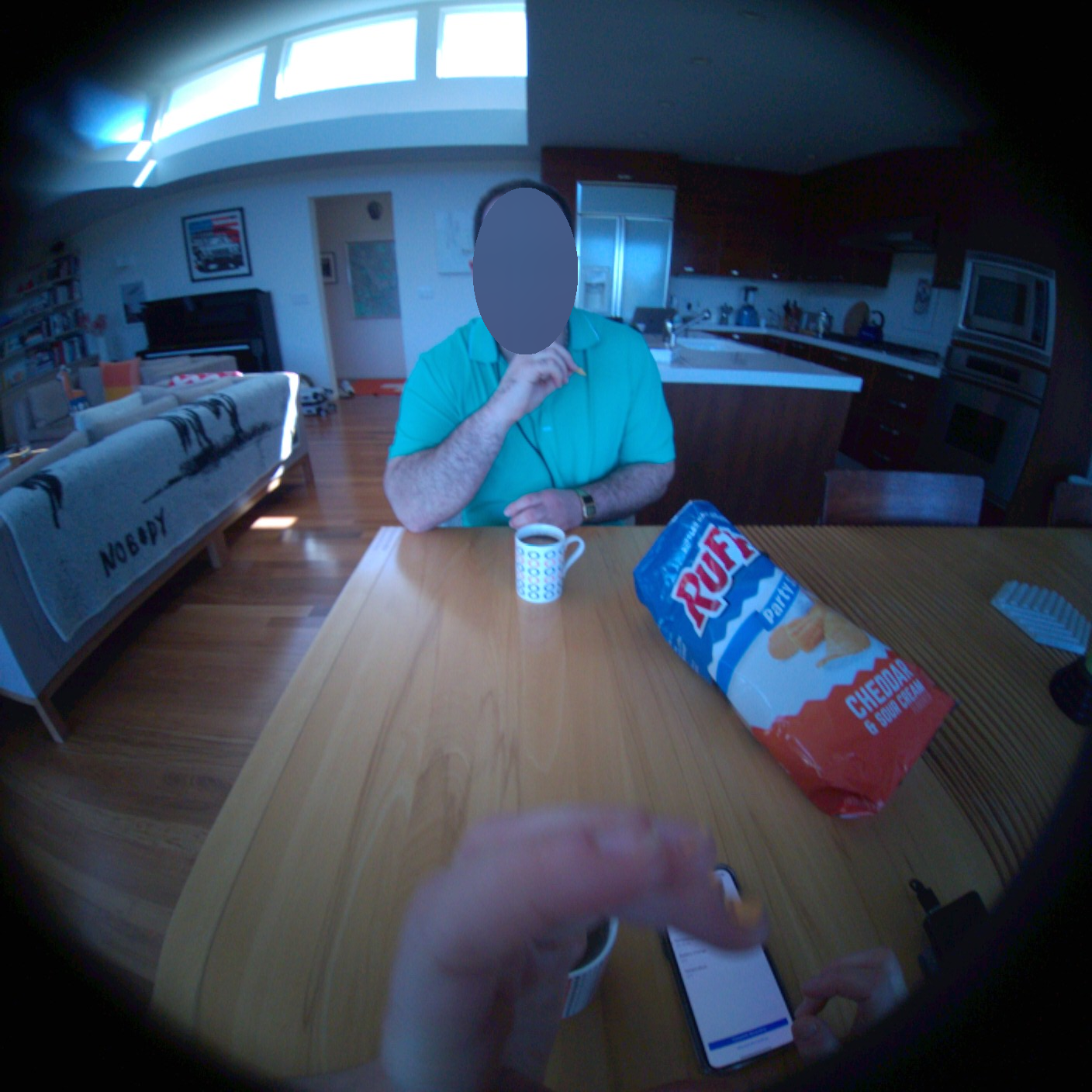}} &
\subfloat[]{\includegraphics[width = 1.5in]{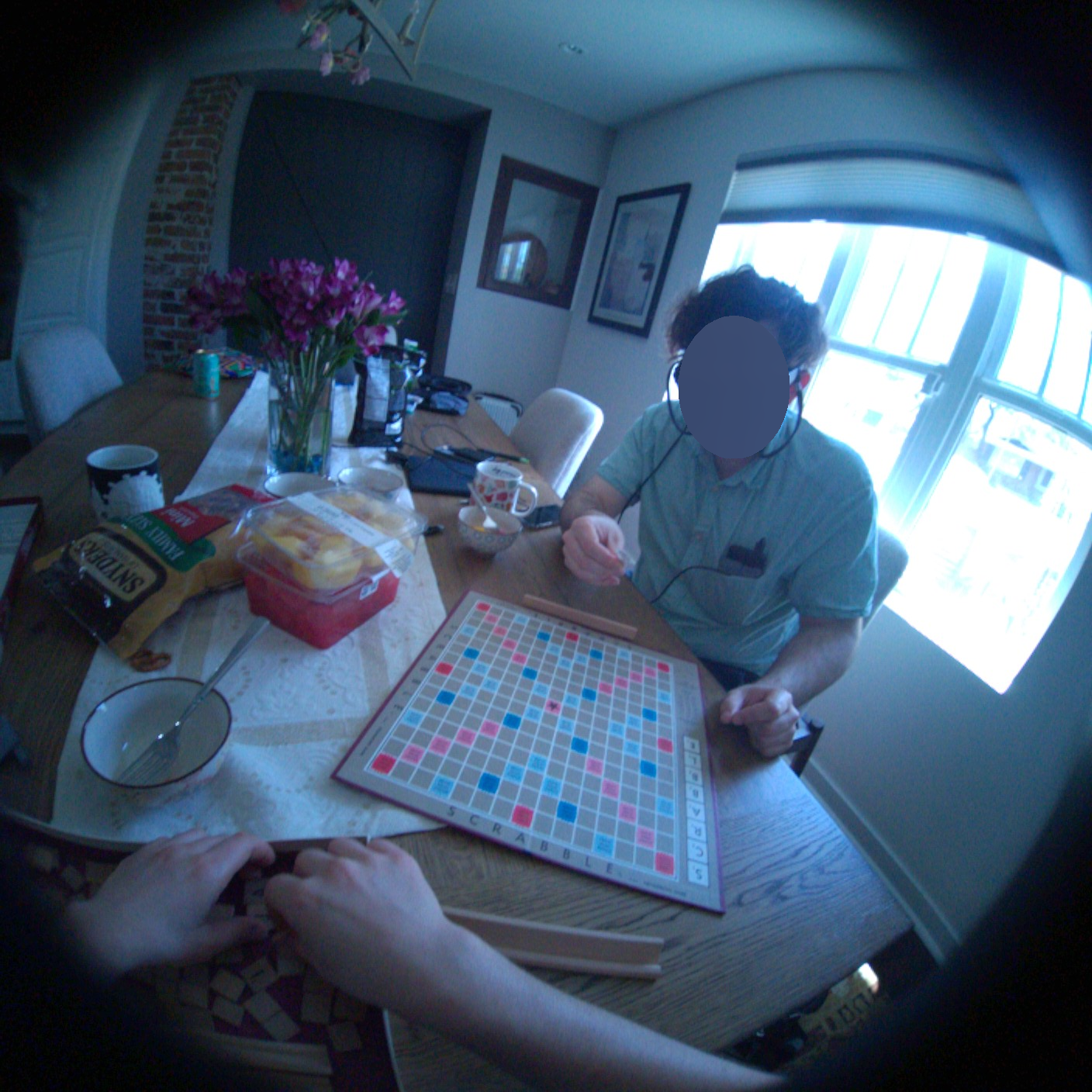}} &
\subfloat[]{\includegraphics[width = 1.5in]{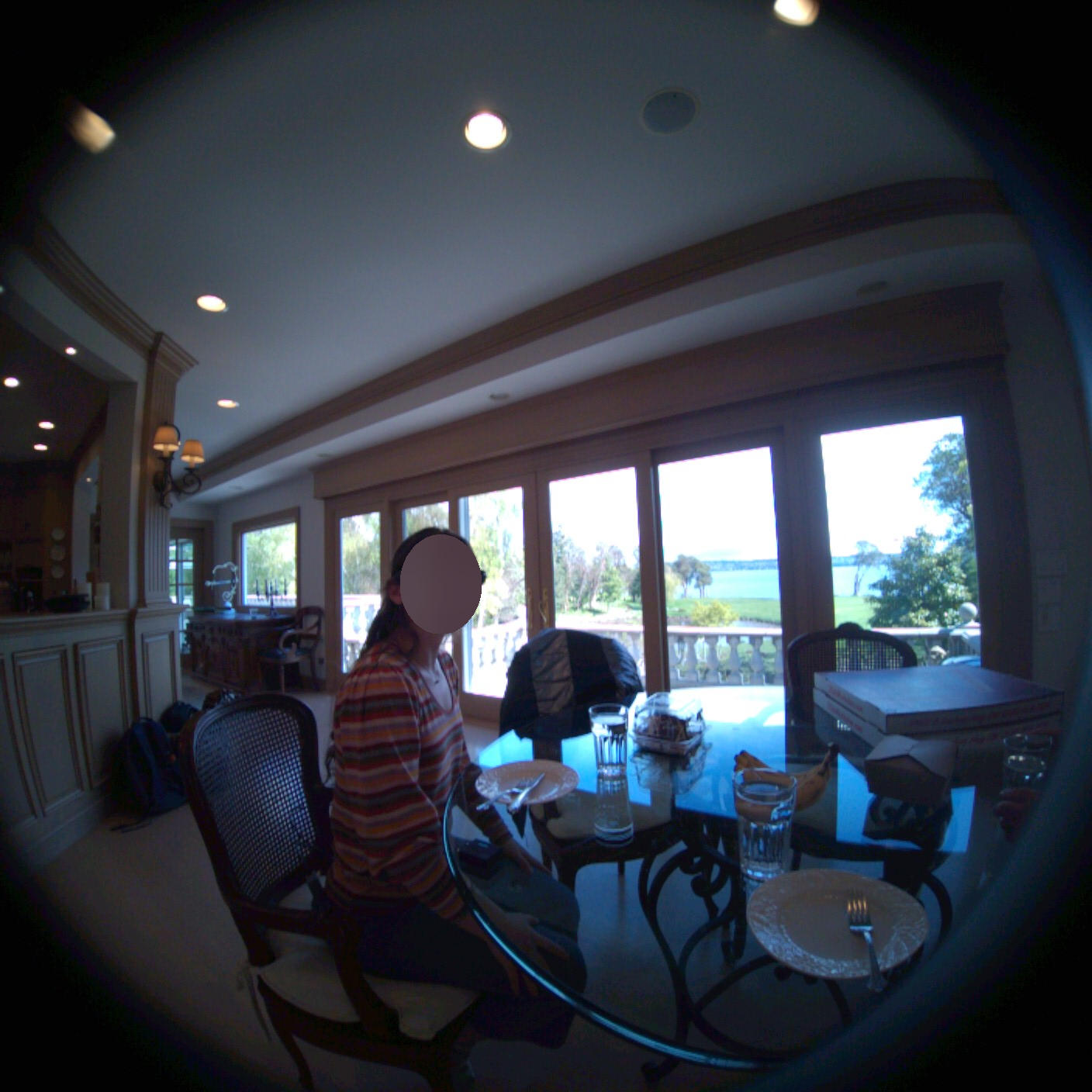}} &
\subfloat[]{\includegraphics[width = 1.5in]{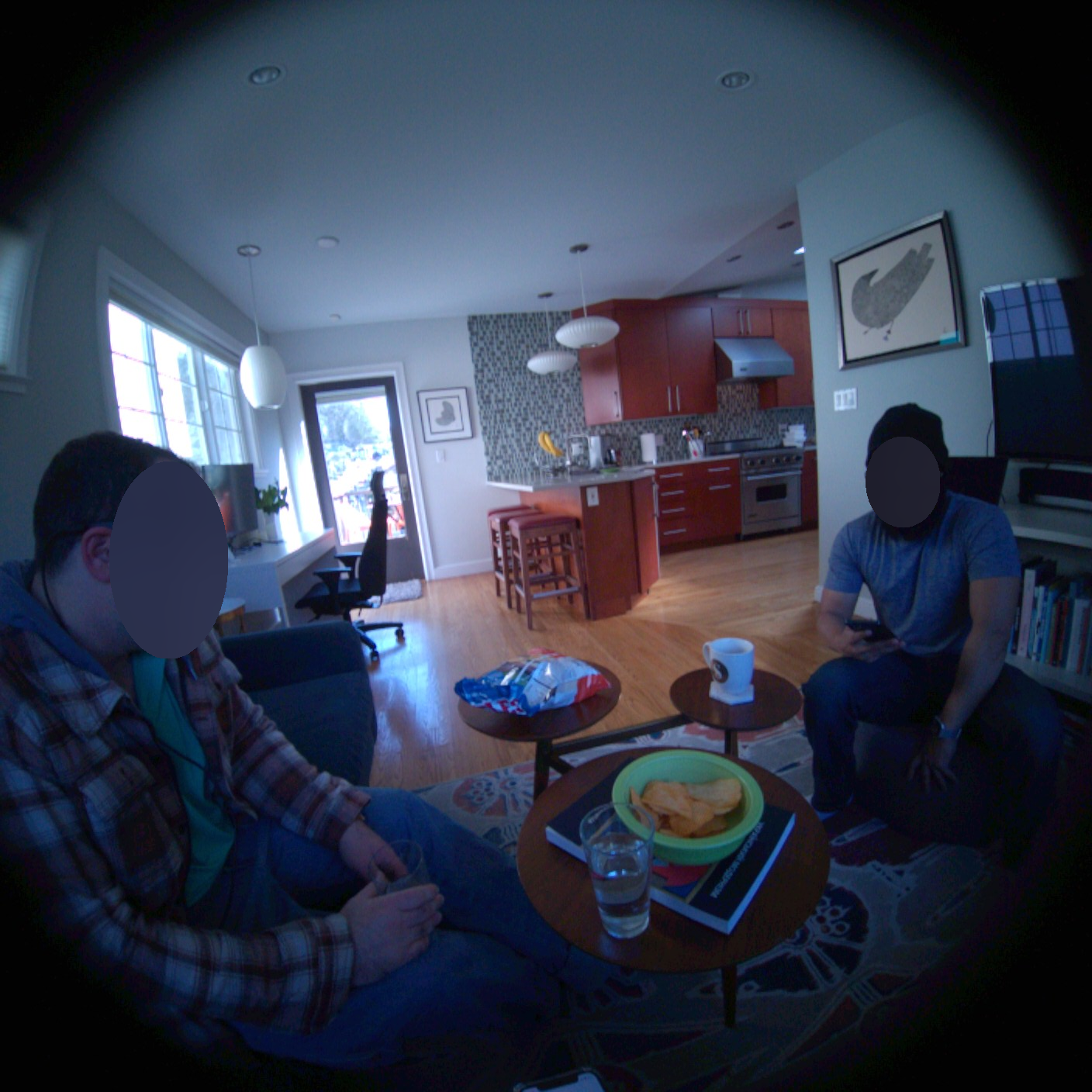}}

\end{tabular}
\caption{Qualitative results on Aria Pilot dataset. Aria Pilot dataset provides challenging benchmark for evaluating performance of our method.}
\label{fig:aria_pilot_face_anon}
\end{figure*}

Tables~\ref{tab:aria_face_rgb_eval} and \ref{tab:aria_face_grayscale_eval} present a more detailed analysis of the performance of all methods on the Aria Pilot Dataset. It can be observed that our method consistently performs at par or better than the state-of-the-art, with an exception for truncated faces. Similar to the observation on CCV2 dataset, the performance of our method is consistent across all Responsible AI buckets. Figure~\ref{fig:aria_pilot_face_anon} showcases some qualitative results of our method on the Aria Pilot dataset.

%Tables~\ref{tab:aria_face_rgb_eval} and \ref{tab:aria_face_grayscale_eval} show the results of more fine grained analysis of all methods, it can be seen that our method consistently performs at par or better than the state of the art. Moreover, with an exception for truncated faces, similar to the observation on CCV2 dataset, performance of our method is consistent across all buckets. Figure~\ref{fig:aria_pilot_face_anon} shows few qualitative results of our method on the Aria Pilot dataset.

\begin{table}[]
\centering
\begin{tabular}{|c|c|c|}
\hline
\textbf {System} & \textbf {Average Precision (AP)} & \textbf {Average Recall (AR)} \\ \hline
\textbf {Mediapipe}    &   0.979       &      0.99    \\ \hline
\textbf {RetinaFace}    &  0.99   &   0.998       \\ \hline
\textbf {EgoBlur}    &   0.99       &   0.998       \\ \hline
\end{tabular}
\caption{Performance comparison of systems using our annotations of CCV2 dataset.}
\label{tab:ccv2_overall_eval}
\end{table}

\begin{table}[]
\centering
\begin{tabular}{|c|c|c|c|}
\hline
\multirow{2}{*}{\textbf {Age}} & \textbf {Mediapipe} & \textbf {RetinaFace} & \textbf {EgoBlur} \\
\cline{2-4}
 & \multicolumn{3}{c|}{\textbf {Average Recall (AR)}}  \\
\hline
\textbf {18-20} & 0.989 & 0.997 & 0.997 \\ \hline
\textbf {20-25} & 0.99 & 0.998 & 0.999  \\ \hline
\textbf {25-30} & 0.99 & 0.998 & 0.998  \\ \hline
\textbf {30-35} & 0.99 & 0.997 & 0.998  \\ \hline
\textbf {35-40} & 0.989 & 0.998 & 0.998  \\ \hline
\textbf {40-45} & 0.99 & 0.998 & 0.998  \\ \hline
\textbf {45-50} & 0.99 & 0.997 & 0.997  \\ \hline
\textbf {50-55} & 0.992 & 0.998 & 0.999  \\ \hline
\textbf {55-60} & 0.986 & 0.997 & 0.998  \\ \hline
\textbf {60-65} & 0.99 & 0.999 & 0.999  \\ \hline
\textbf {65-70} & 0.98 & 0.994 & 0.994  \\ \hline
\textbf {70-75} & 0.983 & 0.992 & 0.992  \\ \hline
\textbf {80-85} & 0.983 & 1 & 1  \\ \hline
\textbf {prefer not to say} & 0.989 & 0.995 & 0.995 \\ \hline
\end{tabular}
\caption{Performance comparison of systems across various age ranges on the CCV2 dataset.}
\label{tab:ccv2_age_eval}
\end{table}

\begin{table}[]
\centering
\begin{tabular}{|c|c|c|c|}
\hline
\multirow{2}{*}{\textbf {Gender}} & \textbf {Mediapipe} & \textbf {Retinaface} & \textbf {EgoBlur} \\
\cline{2-4}
 & \multicolumn{3}{c|}{\textbf {Average Recall (AR)}}  \\
\hline
\textbf {cis woman} & 0.99 & 0.998 & 0.998  \\ \hline
\textbf {cis man} & 0.989 & 0.998 & 0.998  \\ \hline
\textbf {prefer not to say} & 0.996 & 0.997 & 0.997  \\ \hline
\textbf {non-binary} & 0.977 & 0.996 & 0.997  \\ \hline
\textbf {transgender man} & 0.992 & 0.996 & 0.999  \\ \hline
\textbf {transgender woman} & 1 & 1 & 1  \\ \hline
\textbf {None} & 1 & 1 & 1 \\ \hline
\end{tabular}
\caption{Performance comparison of systems across identified genders on the CCV2 dataset.}
\label{tab:ccv2_gender_eval}
\end{table}

\begin{table}[]
\centering
\begin{tabular}{|c|c|c|c|}
\hline
\multirow{2}{*}{\textbf {Fitzpatrick skin tone}} & \textbf {Mediapipe} & \textbf {RetinaFace} & \textbf {EgoBlur} \\
\cline{2-4}
 & \multicolumn{3}{c|}{\textbf {Average Recall (AR)}}  \\
\hline
\textbf {type i} & 0.986 & 0.996 & 0.998  \\ \hline
\textbf {type ii} & 0.989 & 0.997 & 0.998  \\ \hline
\textbf {type iii} & 0.991 & 0.998 & 0.998  \\ \hline
\textbf {type iv} & 0.99 & 0.998 & 0.998  \\ \hline
\textbf {type v} & 0.989 & 0.998 & 0.999  \\ \hline
\textbf {type vi} & 0.98 & 0.997 & 0.997 \\ \hline

\end{tabular}
\caption{Performance comparison of systems across Fitzpatrick skin tone annotations on the CCV2 dataset.}
\label{tab:ccv2_fitzpatrick_skin_tone_eval}
\end{table}

\begin{table}[]
\centering
\begin{tabular}{|c|c|c|c|}
\hline
\multirow{2}{*}{\textbf {Monk skin tone}} & \textbf {Mediapipe} & \textbf {RetinaFace} & \textbf {EgoBlur} \\
\cline{2-4}
 & \multicolumn{3}{c|}{\textbf {Average Recall (AR)}}  \\
\hline
\textbf {scale 1} & 0.987 & 0.995 & 0.998  \\ \hline
\textbf {scale 2} & 0.988 & 0.998 & 0.998  \\ \hline
\textbf {scale 3} & 0.99 & 0.997 & 0.997  \\ \hline
\textbf {scale 4} & 0.99 & 0.998 & 0.998  \\ \hline
\textbf {scale 5} & 0.991 & 0.999 & 0.999  \\ \hline
\textbf {scale 6} & 0.989 & 0.997 & 0.998  \\ \hline
\textbf {scale 7} & 0.99 & 0.996 & 0.997  \\ \hline
\textbf {scale 8} & 0.985 & 0.996 & 0.997  \\ \hline
\textbf {scale 9} & 0.98 & 0.996 & 0.997  \\ \hline
\textbf {scale 10} & 0.966 & 0.997 & 0.997 \\ \hline

\end{tabular}
\caption{Performance comparison of systems across monk scale skin tone annotations on the CCV2 dataset.}
\label{tab:ccv2_monk_skin_tone_eval}

\end{table}

\begin{table*}[]
\centering
\begin{tabular}{|c|c|c|c|c|}
\hline
 \multirow{2}{*}{\textbf {Country}} & \multirow{2}{*}{\textbf {Total Instances}} & \textbf {Mediapipe} & \textbf {RetinaFace} & \textbf {EgoBlur} \\
 \cline{3-5}
 & & \multicolumn{3}{c|}{\textbf {Average Recall (AR)}}  \\
\hline
\textbf {Brazil} & 90210 & 0.989 & 0.997 & 0.997  \\ \hline
\textbf {India} & 69082 & 0.991 & 0.999 & 0.999  \\ \hline
\textbf {Indonesia} & 20809 & 0.989 & 0.998 & 0.997  \\ \hline
\textbf {Mexico} & 19809 & 0.991 & 0.998 & 0.998  \\ \hline
\textbf {Philippines} & 39434 & 0.992 & 0.999 & 0.999  \\ \hline
\textbf {U.S.A} & 19920 & 0.986 & 0.996 & 0.997  \\ \hline
\textbf {Vietnam} & 392 & 0.99 & 0.982 & 0.990 \\ \hline

\end{tabular}
\caption{Performance comparison of systems across subjects of various countries on the CCV2 dataset.}
\label{tab:ccv2_country_eval}
\end{table*}

\begin{table}[]
\centering
\begin{tabular}{|c|c|c|c|c|}
\hline
\multirow{2}{*}{\parbox[c]{1.8cm}{\centering \textbf{System}}} & \multicolumn{2}{c|}{\textbf {Grayscale}} & \multicolumn{2}{c|}{ \textbf{RGB}} \\ \cline{2-5}
                          & \parbox[c]{1.2cm}{\textbf {Average Precision (AP)}} & \parbox[c]{1.2cm}{\textbf {Average Recall (AR)}} & \parbox[c]{1.2cm}{\textbf {Average Precision (AP)}} & \parbox[c]{1.2cm}{\textbf {Average Recall (AR)}} \\ \hline
\parbox[c]{1cm}{\textbf {Mediapipe}}            & 0.203 & 0.39 & 0.426 & 0.533          \\ \hline
\textbf {RetinaFace}                    & 0.806 & 0.825 & 0.877 & 0.905          \\ \hline
\textbf {EgoBlur}                     & 0.866 & 0.899 & 0.895 & 0.938          \\ \hline
\end{tabular}
\caption{Performance of our best system on Aria Pilot dataset compared with Mediapipe face detection system from the industry and RetinaFace detector from academia. Our system achieves over 89.9\% recall on challenging grayscale and RGB egocentric data performing better than the state of the art. }
\label{tab:aria_face_overall_eval}
\end{table}

\begin{table}[]
\centering
\begin{tabular}{|c|c|c|c|}
\hline
\multicolumn{4}{|c|}{\textbf {RGB}} \\ \hline
\multirow{2}{*}{\textbf {Enhanced Labels}} & \multicolumn{3}{c|}{\textbf {Recall}} \\ \cline{2-4}
                          & \textbf {Mediapipe} & \textbf {RetinaFace} & \textbf {EgoBlur} \\ \hline
\textbf {wearing-glasses} & 0.504 & 0.927 & 0.957 \\ \hline
\textbf {non-frontal} & 0.29 & 0.877 & 0.928 \\ \hline
\textbf {truncated} & 0.059 & 0.37 & 0.430 \\ \hline
\textbf {occluded} & 0.359 & 0.851 & 0.892 \\ \hline
\textbf {lighting-too-dark} & 0.109 & 0.818 & 0.854 \\ \hline
\end{tabular}
\caption{Performance comparison of systems on various find grained attributes on RGB (colored) data from the Aria Pilot Dataset. Our solution provides comparable or better performance than the state of the art systems.} 
\label{tab:aria_face_rgb_eval}
\end{table}

\begin{table}[]
\centering
\begin{tabular}{|c|c|c|c|}
\hline
\multicolumn{4}{|c|}{\textbf {Grayscale}} \\ \hline
\multirow{2}{*}{\textbf {Enhanced Labels}} & \multicolumn{3}{c|}{\textbf {Recall}} \\ \cline{2-4}
                          & \textbf {Mediapipe} & \textbf {RetinaFace} & \textbf {EgoBlur} \\ \hline
% \multirowcell{}{\parbox[c]{1.5cm}{\centering System} & \multicolumn{3}{c|}{Recall} \\ \hline
\textbf {wearing-glasses} & 0.319 & 0.841 & 0.914 \\ \hline
\textbf {non-frontal} & 0.256 & 0.805 & 0.880 \\ \hline
\textbf {truncated} & 0.071 & 0.511 & 0.502 \\ \hline
\textbf {occluded} & 0.271 & 0.814 & 0.876 \\ \hline
\textbf {lighting-too-dark} & 0.294 & 0.761 & 0.758 \\ \hline
\end{tabular}
\caption{Performance comparison of systems on various fine grained attributes on Grayscale data from the Aria Pilot Dataset. Our solution provides comparable or better performance than the state of the art systems.}
\label{tab:aria_face_grayscale_eval}
\end{table}

\subsection{License-plates}

Similar to the face detector training described above, for vehicle license plate anonymization, we aimed at establishing a strong baseline performance using the FasterRCNN architecture. Due to the lack of a previous strong baseline model as a strong teacher, we bootstrapped our data engine using training data obtained from manual annotation of large scale images. We created a dataset of over 200K images using this process. Similar to face detector training described above, we used these images for training the FasterRCNN based detector based on the ResNext101-32x8 backbone. 

\paragraph{Benchmarking Dataset}

To benchmark the performance, we collected a comprehensive test dataset using Aria devices. Our in-house data collection team acquired over 40 recordings spanning two weeks at the parking lots of our offices. These recordings were captured under varying conditions such as different times of day, viewing distances, angles, car types, and motion types. We sampled a total of 56,561 frames from these videos and sent them through two phases of manual annotations similar to those performed on the Aria pilot test dataset. The first phase involved labeling boxes, while the second focused on fine-grained attribute annotations.

%To benchmark the performance, we collected a comprehensive test dataset using Aria devices. To construct this dataset, using our in-house data collection team, we acquired over 40 recordings spanning over 2 weeks at the parking lots of our offices. These recordings were captured under varying conditions such as different times of the day, at various viewing distances, viewing angles, car types, motion types, etc.  We sampled a total of 56,561 frames from these videos and those were sent through the two phases of manual annotations similar to the manual annotations performed for Aria pilot test dataset once for labelling boxes and then for fine grained attribute annotations.

\paragraph{Results}
To evaluate the performance of our vehicle license plate anonymization method, we used Intersection Over Union (IoU) with a threshold of 0.5 and average precision and recall as metrics. The results are presented in Table~\ref{tab:aria_lp_overall_eval}, demonstrating strong and consistent performance across both RGB and grayscale streams of the Aria recordings.

%Similar to the evaluation of Face detection method, we used Intersection Over Union with threshold of 0.5 and average precision, average recall as metrics for benchmarking. Table~\ref{tab:aria_lp_overall_eval} shows results of our system measured on the benchmarking dataset. We show strong and consistent performance across both RGB and Grayscale streams of the Aria recordings.

\subsection{Conclusion}

We have successfully developed EgoBlur, a system for face and license plate anonymization in Aria recordings, demonstrating our commitment to preserving the privacy of individuals. Our analysis shows that the face model perform similarly or better than strong baseline methods from academia and industry. The fine-grained performance of this model on Responsible AI datasets is consistent across different buckets and recording streams. Additionally, our analysis provides guidance for future improvements, particularly in anonymizing truncated faces. We also establish a strong baseline model for vehicle license plate anonymization. It's important to note that these models are only trained to locate faces and license plates in images and do not produce any additional attributes. Fine-grained annotations were provided on test data but were not used in training our models.

%We have successfully introduced strong models for face and license plate anonymization in Aria recordings. This is consistent with our strong commitment to preserving privacy of individuals. Our analysis shows that our models perform similarly or better than the strong baseline methods from the academia and industry. Fine grained performance of our method on Responsible AI datasets is consistent across different buckets and recording streams. Additionally, our analysis also provides guidance for the future improvements, mainly for anonymizing the truncated faces. We also establish a strong baseline method for anonymizing vehicle license plates. It is worth noting that these models are only trained to provide the location of faces and licenseplates in a given image and are not capable of producing any additional attributes. Responsible AI annotations and other fine-grained annotations were only provided on the test data and they were not used in the training of our models in any way.

\begin{table}[]
\centering
\begin{tabular}{|c|c|c|c|c|}
\hline
\parbox[c]{1.5cm}{\centering \textbf{System}} & \multicolumn{2}{c|}{\textbf {Grayscale}} & \multicolumn{2}{c|}{ \textbf{RGB}} \\ \cline{2-5}
                          & \parbox[c]{1.2cm}{\textbf {Average Precision (AP)}} & \parbox[c]{1.2cm}{\textbf {Average Recall (AR)}} & \parbox[c]{1.2cm}{\textbf {Average Precision (AP)}} & \parbox[c]{1.2cm}{\textbf {Average Recall (AR)}} \\ \hline
\textbf {EgoBlur}                     & 0.963 & 0.982 & 0.929 & 0.992         \\ \hline
\end{tabular}
\caption{ Performance of our best system on Aria Pilot LP dataset. }
\label{tab:aria_lp_overall_eval}
\end{table}

%%%%%%%%%%%%%%%%%%%%%%%%%%%%%%%%%%%%%%%%%%%%%%%%%%%%%%%%%%%%%%%%%%%%%%%%%%%%%%%%

%%%%%%%%%%%%%%%%%%%%%%%%%%%%%%%%%%%%%%%%%%%%%%%%%%%%%%%%%%%%%%%%%%%%%%%%%%%%%%%%

{\small
\bibliographystyle{ieee}
\bibliography{egbib}

}

\end{document}